%% file: wileyNJD-AMA.tex
\documentclass[AMA,STIX1COL]{WileyNJD-v2}

\input{settings.tex}
\articletype{Research Article}%

\received{17 July 2023}
\revised{Day Month 2023}
\accepted{Day Month 2023}

\begin{document}

\title{Robust Explicit Model Predictive Control for Hybrid Linear Systems with Parameter Uncertainties}

\author[1]{Oleg Balakhnov}

\author[2]{Sergei Savin*}

\author[3]{Alexandr Klimchik}

\authormark{OLEG BALAKHNOV \textsc{et al}}

\address[1]{\orgname{Ocado Technology}, \orgaddress{\state{Hertfordshire}, \country{United Kingdom}}}

\address[2]{\orgdiv{Faculty of Computer Science and Engineering}, \orgname{Innopolis University}, \orgaddress{\state{Tatarstan}, \country{Russia}}}

\address[3]{\orgdiv{School of Computer Science}, \orgname{University of Lincoln}, \orgaddress{\state{Lincolnshire}, \country{United Kingdom}}}

\corres{*Sergei Savin, Innopolis 420500, Tatarstan, Russia. \email{s.savin@innopolis.ru}}


\abstract[Summary]{
\input{sections/abstract}
}


\jnlcitation{\cname{%
\author{O. Balakhnov},
\author{S. Savin}, and
\author{A. Klimchik}} (\cyear{2023}),
\ctitle{Robust explicit model predictive control for hybrid linear systems with parameter uncertainties}, \cjournal{International Journal of Robust and Nonlinear Control}, \cvol{2023;00:1--6}.}

\maketitle


\input{sections/sect_1_intro}

    \input{sections/sect_1_StateOfTheArt}
    \input{sections/sect_2_prelim}

    \input{sections/sect_3_problem}
    \input{sections/sect_4_Method}

\input{sections/sect_5_pendulum}

    \input{sections/sect_6_pendubot}
    \input{sections/sect_7_order_reduction}
    \input{sections/sect_8_conc}

\subsection*{Author contributions}

Oleg Balakhnov proposed the original methods, and conducted experiments; Sergei Savin developed the OCP formulation, provided mathematical formulations of the proposed methods, performed analysis, and wrote the manuscript; Alexandr Klimchik supervised the project, and edited the manuscript.

\subsection*{Financial disclosure}

None reported.

\subsection*{Conflict of interest}

The authors declare no potential conflict of interest.

\clearpage

\section*{Author Biography}

\begin{biography}{\includegraphics[width=66pt,height=86pt,draft]{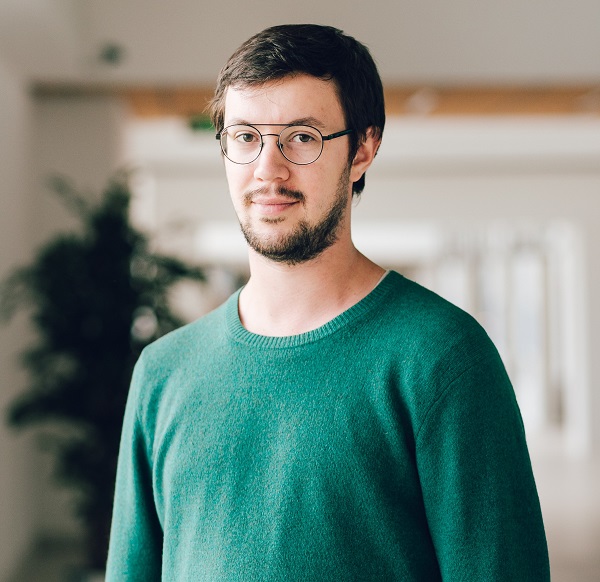}}{\textbf{Oleg Balakhnov} received B.S. degree in radio engineering from Peter the Great St. Petersburg Polytechnic University and M.S. degree in computer science from Innopolis University. 

From 2020 to 2021 was junior researcher at the Center for Technologies in Robotics and Mechatronics Components, Innopolis University, from 2021 to 2023 was working at Control lab, Sber Automotive Technologies. Since 2023 is a Mechatronics Engineer at Ocado Technology (Hatfield, United Kingdom).

His research interest includes underactuated robotics, robust control design, and model predictive control. He did research on  self-driving vehicles, walking robotics, tensegrity systems, variable stiffness, and twisted string actuators.}
\end{biography}

\begin{biography}{\includegraphics[width=66pt,height=86pt,draft]{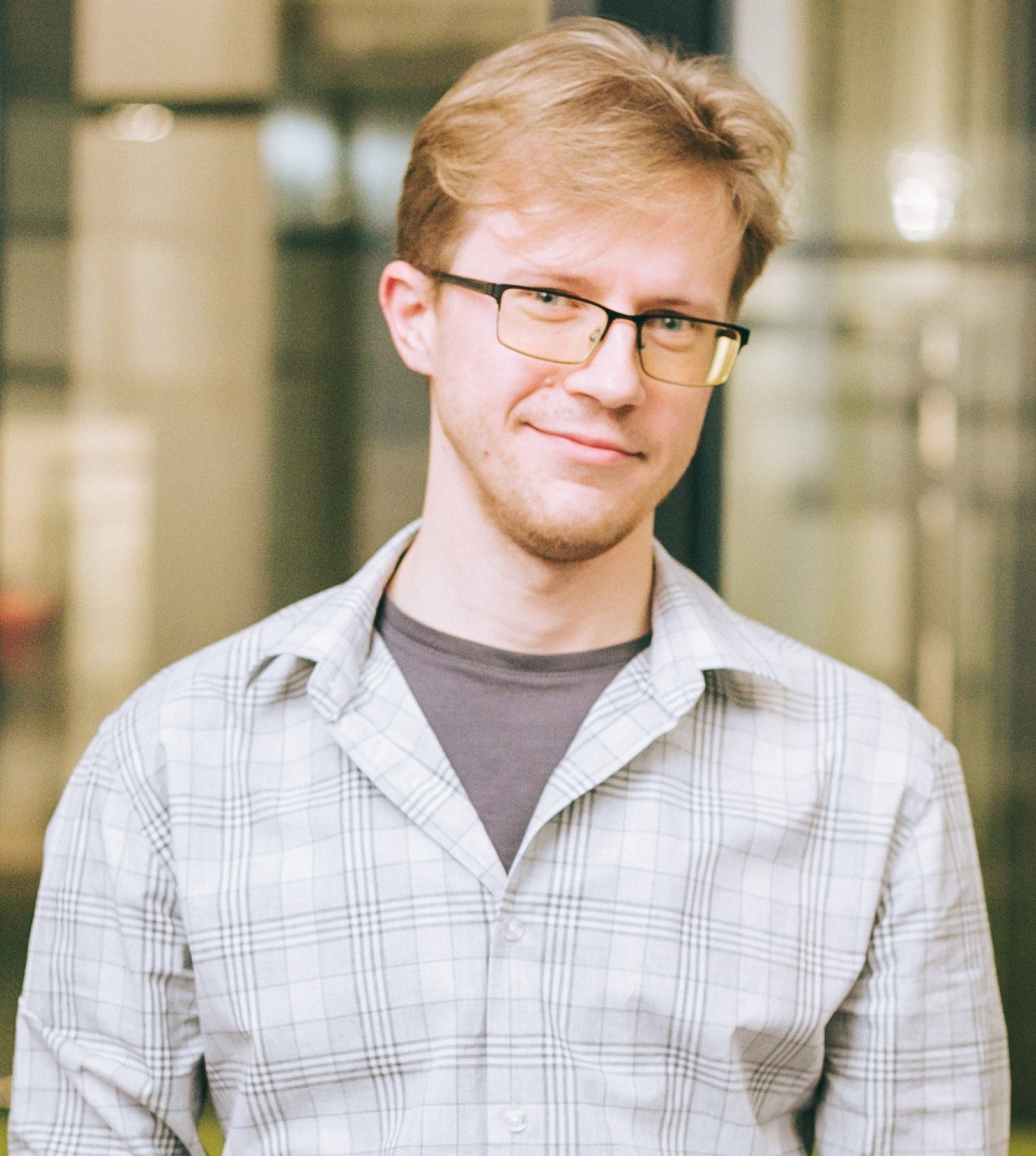}}{\textbf{Sergei Savin} received Ph.D. degree in Mechanical Engineering in 2014 from Southwest State University. Worked as a Teaching Assistant Lecturer, Lecturer, Senior Lecturer, and Docent at the department of Mechanics, Mechatronics and Robotics, Southwest State University from 2013-2018, as a Senior Researcher at Center for Technologies in Robotics and Mechatronics Components, Innopolis University 2018-2022. Since 2019 is an Assistant Professor at the Robotics Institute, Innopolis University.

He is the author of more than 70 papers and 20 patents. His research interests include simulation and control of walking robots, tensegrity structures, in-pipe robots, exoskeletons, multi-link mechanisms, variable stiffness actuators, trajectory optimization, motion over uneven terrain, optimization-based control, use of neural networks in feedback control design and geometric methods in control and state estimation.}
\end{biography}

\begin{biography}{\includegraphics[width=66pt,height=86pt,draft]{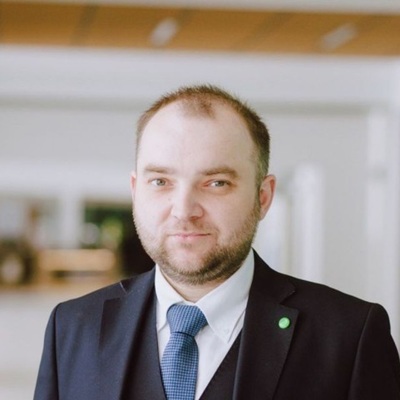}}{\textbf{Alexandr Klimchik} received the Engineering and Master degree in automation control from Belarusian State University of Informatics and Radioelectronics (Belarus) in 2006 and 2007, respectively, Ph.D. degrees in mechanical engineering (robotics) from the Ecole Centrale de Nantes (France) in 2011. From 2006 to 2008 he was a Research Assistant at Belarusian State University of Informatics and Radioelectronics. From 2008 to 2015 he was a Researcher at Ecole des Mines de Nantes (France) and was a member of the Robotics Team in the Research Institute in Communications and Cybernetics of Nantes (IRCCyN, France). From 2015-2022 worked as Assistant and Associate Professor at Innopolis University. Since 2022 is Associate Professor in Robotics at University of Lincoln.

His research covers stiffness modeling, robot calibration, and human-robot collaboration; he focused on stiffness modeling of fully-actuated, over-constrained, and under-constrained manipulators, as well as on control and error compensation for robots with direct and indirect feedback control. The obtained theoretical results have been applied to stiffness analysis of industrial robots of serial, quasi-serial, and parallel architecture, medical robots, humanoid robots, cable-driven robots, and collaborative robots. These contributions were published in more than 120 scientific papers; 7 journal papers were among Top 10 most cited articles in the leading journals during the year (MaMT, RCIM, FINEL).}
\end{biography}

\end{document}

%% file: settings.tex
\newcommand{\norm}[1] {\left|\left|{#1}\right|\right|}
\newcommand{\Zo}[2] {\left\langle {#1},{#2}
\right\rangle}

\newtheorem{example}{Example}

%% file: sections/abstract.tex
Explicit model-predictive control (MPC) is a widely used control design method that employs optimization tools to find control policies offline; commonly it is posed as a semi-definite program (SDP) or as a mixed-integer SDP in the case of hybrid systems. However, mixed-integer SDPs are computationally expensive, motivating alternative formulations, such as zonotope-based MPC (zonotopes are a special type of symmetric polytopes). In this paper, we propose a robust explicit MPC method applicable to hybrid systems. More precisely, we extend existing zonotope-based MPC methods to account for multiplicative parametric uncertainty. Additionally, we propose a convex zonotope order reduction method that takes advantage of the iterative structure of the zonotope propagation problem to promote diagonal blocks in the zonotope generators and lower the number of decision variables. Finally, we developed a quasi-time-free policy choice algorithm, allowing the system to start from any point on the trajectory and avoid chattering associated with discrete switching of linear control policies based on the current state's membership in state-space regions. Last but not least, we verify the validity of the proposed methods on two experimental setups, varying physical parameters between experiments. 
%
%
%
%
%
%

%% file: sections/sect_1_intro.tex
\section{Introduction}

Model-based control is an essential tool in many areas of engineering, especially in Robotics. Its general weakness relates to our limited ability to exactly model the plant; examples of that are parametric uncertainty, unmodeled dynamics, unmodeled disturbances, etc. When a sufficiently precise model is not available the model uncertainty needs to be explicitly addressed in a model-based control framework, for which families of methods have been designed. Examples of the latter include adaptive and robust control \cite{ortega2019adaptive, karimi2007robust}; the latter is well represented by linear-matrix inequality (LMI)-based methods for systems with polytopic, interval, and norm-bounded uncertainties \cite{karimi2007robust, rosinova2003robust, amato2011robust}, as well as for systems with unknown inputs, and others. 

Among the model-based methods that have seen a lot of progress in the last three decades is control design based on set propagation. Advantages of this type of control design include the ability to directly account for the set of initial conditions and to place constraints on the final and intermediate sets of control actions, states, and/or outputs. Limiting ourselves to two examples of control design, we can mention the propagation of ellipsoidal sets employed in Finite-Time Stability (FTS) control \cite{amato2014finite} and propagation of zonotopes employed in reachability analysis and recently in control design for piecewise affine (PWA) dynamical systems \cite{sadraddini2019sampling}. Zonotopes are symmetric polytopes, described as an affine transformation of a unit cube; the linear part of the transformation is called \emph{generator}, and the additive part is called \emph{center}. Their properties will be discussed in the later sections. Both types of set propagation allow casting control design problem as a single convex optimization program, which is one of the chief appeals of these approaches. In the case of linear dynamical systems, propagation of ellipsoidal sets in FTS control usually leads to linear matrix inequalities (LMI) \cite{dorato1997robust, amato2006finite, amato2014finite} and hence semidefinite programs (SDP) \cite{boyd1994linear}, while the propagation of zonotopes can lead to linear programs (LP), quadratic programs (QP) or second-order cone programs (SOCP), depending on the objective function and set containment criteria used \cite{sadraddini2019sampling}. For piecewise affine dynamical systems, control design requires integer variables which leads to mixed-integer programs. Among mentioned types of optimization problems, SDP is the most challenging, having less mature solvers compared with SOCP, QP, and LP. This is especially apparent with mixed-integer problems \cite{sadraddini2019sampling}. This justifies the interest in control design based on zonotope propagation.

Robust control design methods for FTS and are well developed, taking advantage of the results in LMI-based control for linear systems \cite{dorato1997robust, amato2014finite}. The latter offer methods for handling norm-bounded uncertainties, polytopic and interval uncertainties, and unknown inputs, casting all of those problems as a single LMI \cite{kothare1996robust, amato2006robust, amato2011robust, li2008improved, ramos2002lmi}. The same methods are currently lacking for control design based on the propagation of zonotopes. We should note that this statement refers only to feedback control design, as robust state estimation methods based on the use of zonotopes for set representation have been studied previously \cite{combastel2015zonotopes, wang2018set}. It is also limited to zonotope propagation cast as a set of linear transformations and Minkowski sums of zonotopes presented as centers and generators, which leads to linear, quadratic, and second-order cone programs; there are works that implement, e.g., stabilization of systems with uncertainties based on invariant sets represented as zonotopes, where computations are cast in the traditional LMI framework leading to SDP problems \cite{hamdi2017stabilization}. The goal of this work is to provide a control design framework based on zonotope propagation, robust to additive and polytopic multiplicative model uncertainties in both the state and control matrices.

An important limitation, characteristic of a number of robust control methods for linear systems with multiplicative time-varying uncertainties is that the set of all possible uncertain parameters is itself time-invariant. Examples of control methods developed for such systems include the aforementioned LMI-based robust control for LTI systems, as well as a number of robust control methods for linear parameter-varying (LPV) systems \cite{apkarian1995self, amato2006robust}. For models obtained as a linearization of an uncertain nonlinear system along a nominal trajectory, the set of uncertain parameters of the resulting model may itself be time-variant. We show that this type of model can be handled in the proposed framework; the proposed method can also be extended to hybrid linear dynamical systems, as we show in the paper.

Thus, this paper proposes a robust explicit MPC method for time-varying linear and hybrid linear dynamical systems based on the propagation of zonotopes, allowing to directly handle time-varying polytopic multiplicative uncertainties, as well as additive uncertainties and constraints on state and control. To facilitate the practical application of the proposed method, we introduce a new zonotope order reduction method, as well as a time-agnostic approach to feedback control. To the best of our knowledge, this is the first time a zonotope set propagation-based explicit MPC robust to time-varying polytopic parametric uncertainty was proposed, and the first time a zonotope-based robust MPC was experimentally validated.

%% file: sections/sect_1_StateOfTheArt.tex
\section{State of the Art}

Zonotope set representation has been successfully used in various areas control: in reachability analysis \cite{alanwar2021data}, formal system verification \cite{althoff2008verification}, control design based on invariant sets \cite{wan2009numerical, hamdi2017stabilization, han2016enlarging}, fault detection \cite{xu2013relationship, yang2020novel}, and state estimation \cite{alamo2005guaranteed, le2013zonotopic}, among other applications. The questions of robustness of these methods to model uncertainties and uncertain inputs have been studied as well \cite{le2013zonotopic, combastel2015zonotopes}.

Recently, a control design for piecewise affine systems was proposed based on zonotope set propagation \cite{sadraddini2019sampling}. It can be seen as similar to explicit MPC resulting from considering finite-time stability of a time-varying linear system with ellipsoidal state sets \cite{amato2014finite}. There are however a number of differences between the two approaches. The primal difference is that the FTS-based control design makes use of Lyapunov equations and casts the problem as an SDP, whereas the zonotope propagation method directly applies dynamics equations to the state set, casting the problem as an LP, QP, or SOCP. The latter has many advantages but lacks techniques for dealing with multiplicative model uncertainty; relevant LMI-based methods do not have zonotope-based analogs. This paper aims to alleviate part of this problem by proposing a control design method robust to polytopic model uncertainties.

There are a number of classical results in robust control of systems with norm-bounded and polytopic uncertainty \cite{amato2006robust}. Quadratic stability of a linear time-invariant (LTI) system with polytopic model uncertainty (where the set of all possible models is a polytope) can be proven by simultaneously solving Lyapunov equations for the vertices of the polytopic set of possible models, which can be cast as a single LMI \cite{amato2006robust}; this result allows to design robust control both for uncertain LTI and linear parameter varying systems and can be naturally extended to control design. For systems with norm-bounded model uncertainty, the problem can be reformulated as convex using linear-fractional transformation and S-procedure, also enabling control design as a single LMI \cite{amato2014finite}. This approach can be extended to include input and output constraints, leading to a robust explicit MPC formulation \cite{de2004robust}. While these results have vast practical significance, the appeal of zonotope propagation-based techniques is the lower computational complexity, which gives them the ability to handle hybrid systems while using mixed-integer solvers as a back-end \cite{sadraddini2019sampling}, as well as the possibility to directly account for additive disturbance using Minkowski sum of zonotopes.

Control design based on zonotope set propagation faces a number of well-known problems associated with numerical operations on zonotopes in general, and with the use of Minkowski sums in particular. One of these problems is zonotope containment. There are a number of works on the topic, with an array of algorithms proposed \cite{han2016enlarging, sadraddini2019linear, kulmburg2021co}. However, since the control design is required to be cast as convex optimization, the range of possible zonotope containment algorithms is limited to the ones that can be presented as linear or SOCP problems. Such an algorithm has been presented in \cite{sadraddini2019linear}. A critical study of this encoding can be found in \cite{kulmburg2021co}.

Another problem is zonotope order reduction methods. The result of Minkowski summation of two zonotopes is a zonotope with a larger generator; this means that iterative application of Minkowski sums leads to a linear increase in the number of elements in the zonotope generators, which in turn means a linear increase in the number of continuous variables in the resulting optimization problem. It has been observed that replacing a zonotope with a new one whose generator has fewer columns leads to a better performance in terms of zonotope propagation and control design \cite{sadraddini2019sampling}. A number of order reduction methods have been proposed \cite{althoff2015introduction, girard2005reachability, combastel2003state, althoff2010reachability, scott2016constrained}. However, those methods are not optimized for an iterative application. In this paper, we exploit the iterative nature of the state propagation problem to propose a novel order reduction method which 1) facilitates the diagonal blocks in zonotopes, 2) prevents the increase of zonotope order after Minkowski summation for a certain class of disturbances.

Thus, the main contributions of the paper are the following:

\begin{itemize}
    \item An extension of the existing zonotope-based explicit MPC to cover the case of time-varying polytopic parametric uncertainty.
    \item A novel convex zonotope order reduction method, that takes advantage of the iterative structure of the convex program which implements the zonotope-based explicit MPC.
    \item A parallelotope-based time-free control policy choice algorithm that solves the time initialization problem (i.e., the problem of starting the motion from an arbitrary part of the trajectory, rather than from its beginning) and avoids chattering.
\end{itemize}

%% file: sections/sect_2_prelim.tex
\section{Notation and Preliminaries}
A zonotope can be defined as a point-symmetric set in $n$-dimensional space \cite{kopetzki2017methods}, described as a center $c \in \mathbb{R}^n $ and $p$ generators $g^{(i)} \in \mathbb{R}^n, \ i \in \{1,...,p\}$; the latter can be presented as columns of matrix $G \in \mathbb{R}^{n \times p}$:
\begin{equation} 
\label{eq_zonotope_def}
    \mathbb{Z} = \Zo{c}{G} = \{c + \sum_{i=1}^p \beta_i g^{(i)} :  -1 \leq \beta_i \leq 1 \}
\end{equation}
where $\beta_i$ are scalar multipliers. Eq. \eqref{eq_zonotope_def} defines a \emph{vector zonotope} (it represents a set whose elements are vectors), as oppose to a \emph{matrix zonotope} which we introduce later. 
In the following discussion, we will also use the concept of \emph{zonotope order}, which is a ratio between the number of columns and rows of the generator. The order of zonotope \eqref{eq_zonotope_def} is given as $p/n$.

\subsection{Algebraic Operations on Vector Zonotopes}
Vector zonotopes are closed under addition and linear transformation, defined as follows:
\begin{equation} 
\label{eq_zonotope_addition}
    \Zo{x}{X}+\Zo{y}{Y} = \Zo{x+y}{X+Y}
\end{equation}
\begin{equation} 
\label{eq_zonotope_linear_tranform}
    A\Zo{c}{G} = \Zo{Ac}{AG},
\end{equation}
where $A$ is a linear operator.

Minkowski sum for vector zonotopes is defined as:
\begin{equation}
\label{eq:MinkowskiDef}
    \Zo{x}{X} \oplus \Zo{y}{Y} = \Zo{x + y}{(X, Y)}
\end{equation}
where notation $(X, Y)$ refers to horizontal matrix concatenation.

We can define the addition of a vector zonotope and a vector as follows:
\begin{equation} 
\label{eq_zonotope_vector_addition}
    \Zo{c}{G}+v = \Zo{c+v}{G}
\end{equation}

That definition implies the following property of the vector addition and Minkowski sum:
\begin{equation}
\label{eq_single_vector_Minkowski}
    \Zo{x + v}{X} \oplus \Zo{y}{Y} = \Zo{x}{X} \oplus \Zo{y + v}{Y} 
\end{equation}

\subsection{Zonotopes with positive-semidefinite diagonal blocks}

Given zonotopes $\Zo{x}{(D_x, X)}$ and $\Zo{y}{(D_y, Y)}$, where $D_x \succeq 0$ and $D_y \succeq 0$ are diagonal matrices, their Minkowski sum can be described by the following zonotope:
\begin{equation}
\label{diagMinkowski}
    \Zo{x}{(D_x, X)} \oplus \Zo{y}{(D_y, Y)} \equiv
    \Zo{x + y}{(D_x+D_y, X, Y)},
\end{equation}
where equivalence is understood in the sense that the set to the left of the sign $\equiv$ contains all vectors that are contained in the set to the right, and no others. Note that this definition of Minkowski sum results in zonotopes with fever columns in the generator matrix; however, this definition only works for zonotopes with positive semidefinite diagonal blocks in their generators.

\subsection{Zonotope Containtment}
To check if a zonotope is contained in another zonotope, we can use the method proposed in \cite{sadraddini2019linear}.
Given two zonotopes $\mathbb{X}$ = $\langle x,X \rangle$ and $\mathbb{Y}$ = $\langle y,Y \rangle$, where $X \in \mathbb{R}^{n \times n_x}$, $Y \in \mathbb{R}^{n \times n_y}$, if there exists $\Gamma \in \mathbb{R}^{n_y \times n_x}$ and $\beta \in \mathbb{R}^{n_y}$, such that:
\begin{equation}
    \label{eq:containment}
    X = Y\Gamma, \ y - x = Y\beta, \  ||(\Gamma,\beta)||_\infty \leq 1
\end{equation}
then zonotope $\mathbb{X}$ is contained in zonotope $\mathbb{Y}$. This method is convenient, as it requires solving a single linear program.
\subsection{Approximating convex hull of two zonotopes}
In \cite{girard2005reachability} it is proposed to use the following approximation of the convex hull of two zonotopes $\mathbb{X} = \langle x,X \rangle$ and $\mathbb{Y} = \langle y,Y \rangle$:
\begin{equation}
    \label{eq_convhull}
    \text{Co}(\mathbb{X}, \mathbb{Y}) \subseteq 
    \Zo{\frac{x + y}{2}}
    {\left(
    \frac{X + Y}{2},\frac{x - y}{2},\frac{X - Y}{2}
    \right)}
\end{equation}

This method can be used to approximate a convex hull of a set of $2^p$, dividing the set into $2^{p-1}$ pairs and applying the method to each pair, and repeating the same step on the resulting set of zonotopes, iterating $p$ times. 

This method is conservative and computationally inexpensive. Most importantly for our purpose, it can be incorporated in a convex optimization problem formulation as a linear equality constraint. 

\subsection{Matrix zonotopes}
Zonotopes can be used to represent symmetric polytopic sets of matrices; in that case they are referred to as \emph{matrix zonotopes} \cite{althoff2011reachable}. A matrix zonotope is defined analogous to a vector zonotope, as $\mathbb{A} = 
    \Zo{A^{(0)}}{ \{ A^{(1)}, ..., A^{({n_a})} \} }$:
\begin{equation} 
\label{eq_matrix_zonotope_def}
    \mathbb{A} = 
    \{A^{(0)} + \sum_{i=1}^{n_a} \beta_i A^{(i)} :  -1 \leq \forall\beta_i \leq 1 \},
\end{equation}
where $A^{(i)}$ are matrices. One can define multiplication of matrix zonotopes $\mathbb{A} \otimes \mathbb{B}$ as:
\begin{equation}
\begin{aligned}
     \mathbb{A} \otimes \mathbb{B} = \{ AB, \ A \in \mathbb{A}, \ B \in \mathbb{B} \}
\end{aligned}
\end{equation}

In \cite{althoff2011reachable} this operation is defined for sets of square matrices; however, given appropriate dimensions of generator matrices, it can be extended to matrix-vector multiplication without changes in the formulations. Following \cite{althoff2011reachable}, an overapproximation of the matrix zonotope multiplication can be introduced as follows. Given a matrix zonotope \eqref{eq_matrix_zonotope_def} and a vector zonotope \eqref{eq_zonotope_def}, their product is approximated as:
\begin{equation}
\label{eq_matrix_zonotope_mult}
\begin{aligned}
     \mathbb{A} \otimes \mathbb{Z} \ \approx
     \Bigl\langle A^{(0)} c, \  
     \Bigl(
     A^{(0)}g^{(1)}, \ ..., A^{(0)}g^{(n_p)}, \ 
     A^{(1)} c, \ A^{(1)}g^{(1)}, \ ..., A^{(1)}g^{(n_p)}, \ 
     ..., \ 
     A^{(n_a)} c, \ A^{(n_a)}g^{(1)}, \ ..., A^{(n_a)}g^{(n_p)} 
     \Bigr)\Bigr\rangle
\end{aligned}
\end{equation}

Implementation of the operations presented in this section can be found in the software package CORA \cite{althoff2015introduction}.

%% file: sections/sect_3_problem.tex
\section{Problem Formulation}

Consider an uncertain discrete affine time-variant (LTV) system:
\begin{equation}
\label{eq:LTV}
    \begin{matrix}
    x_{k+1}  =  A(k) x_k + B(k) u_k + d(k) + w_k
    \\[0.5em]
    [A(k) \ B(k) \ d(k)] \in \Omega_k
    \\[0.5em]
    \end{matrix}
\end{equation}
where 
$x_k \in \mathbb{R}^n$ is the state of the system, 
$u_k \in \mathbb{R}^m$ is the control input, 
$w_k \in \mathbb{R}^n$ is a bounded process disturbance, 
$A(k) \in \mathbb{R}^{n \times n}$ is state matrix, 
$B(k)\in \mathbb{R}^{n \times m}$ is control matrix, 
$d(k) \in \mathbb{R}^n$ is additive term of the affine dynamical model,
and 
$\Omega_k = \text{convexhull}(\mathcal{V}_k)$, 
is a polytopic set of models, defined as a convex hull of its $L$ vertices $\mathcal{V}_k = \left\{
    [A_1(k) \ B_1(k) \ d_1(k)], ..., 
    [A_L(k) \ B_L(k) \ d_L(k)]
    \right\}$. We call the models in the set $\mathcal{V}_k$ \emph{vertex models}.
    
First, we note that the polytopic set $\Omega_k$ is itself time-varying, which well reflects the scenario when $\Omega_k$ is obtained from a linearization of an uncertain non-linear system along a given trajectory. Second, while both $d(k)$ and $w_k$ appear in the expression \eqref{eq:LTV} as additive terms, we avoid grouping them into a single term, since they belong to different sets and will be handled differently in set propagation in the case of uncertain dynamics. Now we can formulate the problem that this paper aims to solve:
    
\begin{problem}
\label{p:problem1}
    For the system \eqref{eq:LTV} find a trajectory $\bar{u}_k$, $\bar{x}_k$ and control policy $u_k = (\bar{u}_k - K_k (x_k - \bar{x}_k) ) \in \mathbb{H}_k^u$, such that for any initial condition $x_0 \in \mathbb{X}_0$, the intermediate values of the state $x_k$ are bounded by $x_k \in \mathbb{H}_k^x$, for any disturbance $w_k \in \mathbb{W}_k$, and for any $[A \ B \ d] \in \Omega_k$, $\forall k$, where $\mathbb{H}_k^u$, $\mathbb{H}_k^x$, $\mathbb{X}_0$ and $\mathbb{W}_k$ are zonotopes.
\end{problem}

\subsection{Known parameters case}
If the matrices $A$ and $B$ are known exactly, the problem \ref{p:problem1} can be described as an evolution (propagation) of the initial zonotope $\mathbb{X}_0$, subject to dynamics \eqref{eq:LTV}:
\begin{equation}
\label{eq:LTV_zonotopes}
    \mathbb{X}_{k+1}  = (A(k) \mathbb{X}_k + B(k) \mathbb{U}_k + d(k)) \oplus \mathbb{W}_k
\end{equation}
where 
$\mathbb{X}_k = \Zo{\bar{x}_k}{G_k}$, 
$\mathbb{U}_k = \Zo{\bar{u}_k}{\theta_k}$, 
and 
$\mathbb{W}_k = \Zo{0_{n \times 1}}{W(k)}$ are zonotopes representing state, control actions, and process noise, 
$G_k      \in \mathbb{R}^{n \times p}$, 
$\theta_k \in \mathbb{R}^{m \times p}$, and 
$W(k)      \in \mathbb{R}^{n \times n_w}$ are generators or these zonotopes, 
$\bar{x}_k      \in \mathbb{R}^{n}$ and 
$\bar{u}_k \in \mathbb{R}^{m}$ 
are their centers. 
Using \eqref{eq_single_vector_Minkowski} we could combine $ d(k)$ and $\mathbb{W}_k$ into a single zonotope $\Zo{d(k)}{W(k)}$, arriving at a standard linear dynamics representation. On each time step any admissible disturbance $w_k \in \mathbb{W}_k$ can act on the system, which is the reason for the use of Minkowski sum; therefore the order of the zonotopes $\mathbb{X}_k$ grows on each time step unless order reduction techniques are employed. 

As it was discussed in \cite{sadraddini2019sampling, sadraddini2019linear}, propagation of zonotopes can be decomposed into separate equations describing the evolution of their centers and generators. With that \eqref{eq:LTV_zonotopes} can be re-written as:
\begin{equation}
\label{eq:generator_prop}
    \begin{cases}
    G_{k+1} = (A(k) G_{k} + B(k) \theta_k, \ W(k)) \\
    \bar{x}_{k+1} = A(k) \bar{x}_{k} + B(k) \bar{u}_k + \bar{d}(k) 
    \end{cases}
\end{equation}

For the case when $A$ and $B$ are known exactly, the following linear control law was proposed in \cite{sadraddini2019sampling}:
\begin{equation}
\label{eq:lin_feedback}
    u = \bar{u}_k - \theta_k G_k^\dagger(x_k - \bar{x}_k)
\end{equation}
where $(\cdot)^\dagger$ denotes Moore-Penrose pseudoinverse. However, the problem \ref{p:problem1} does not allow precise knowledge of $A$ and $B$; our solution to this problem is discussed in the next section.

%% file: sections/sect_4_Method.tex
\section{Robust Constrained Explicit MPC for Hybrid Linear Systems with Parameter Uncertainties}

\subsection{Robustness to parametric uncertainty}

Given an uncertain dynamical system \eqref{eq:LTV} and zonotopes $\mathbb{X}_{k}$ and $\mathbb{U}_{k}$, 
let $\mathbb{X}_{k+1}$ be a zonotope that contains all $x_{k+1}$ that can be obtained by applying \eqref{eq:LTV} to $x_k \in \mathbb{X}_{k}$ and $u_k \in \mathbb{U}_{k}$:
\begin{equation}
\label{eq:LTV_zonotopes_robust}
    \mathbb{X}_{k+1} \supset 
    \bigcup\limits_{[A,B,d] \in \Omega_k} 
    \left( (A(k) \mathbb{X}_k + B(k) \mathbb{U}_k + d(k)) \oplus \mathbb{W}_k \right)
\end{equation}

Expression \eqref{eq:LTV_zonotopes_robust} cannot be directly included in a convex optimization procedure. In order to make the problem numerically tractable, we introduce the following relaxation:
\begin{equation}
\label{eq:LTV_zonotopes_robust_vert}
    \mathbb{X}_{k+1} \supset 
    \bigcup\limits_{\mathcal{V}_k} 
    \left( (A_i(k) \mathbb{X}_k + B_i(k) \mathbb{U}_k + d_i(k)) \oplus \mathbb{W}_k \right)
\end{equation}

Proposed relaxation can be interpreted as follows: instead of searching for a zonotope that contains all possible transformations $A(k) \mathbb{X}_k + B(k) \mathbb{U}_k + d(k)$ for $[A(k),B(k),d(k)] \in \Omega_k$ we limit it to transformations $[A_i(k),B_i(k),d_i(k)] \in \mathcal{V}_k$, i.e. the vertices of $\Omega_k$.

Let us denote $\mathbb{Z}_{k, i} = A_i(k) \mathbb{X}_k + B_i(k) \mathbb{U}_k + d_i(k)$. Let us observe that the fact that \eqref{eq_convhull} is an over-approximation implies $\mathbb{Z}_{k, i} \subset \underset{\mathcal{V}_k}{\text{Co}}(\mathbb{Z}_{k, i})$, which in turn implies:
\begin{equation}
\label{eq_minkowski_of_hull}
    \forall \mathbb{Z}_{k, i} \oplus \mathbb{W}_k \subset \underset{\mathcal{V}_k}{\text{Co}}(\mathbb{Z}_{k, i}) \oplus \mathbb{W}_k,
\end{equation}
where $\underset{\mathcal{V}_k}{\text{Co}}(\cdot)$ means that convex hull is taken over all $\mathbb{Z}_{k, i}$ that can be formed with $[A_i(k), B_i(k), d_i(k)] \in \mathcal{V}_k$. With that we can find an over-approximation of the union in \eqref{eq:LTV_zonotopes_robust_vert}:

\begin{equation}
\label{eq:LTV_zonotopes_robust_hull}
    \bigcup\limits_{\mathcal{V}_k} 
    \left( \mathbb{Z}_{k, i} \oplus \mathbb{W}_k \right) 
    \subset
    \underset{\mathcal{V}_k}{\text{Co}}
    (\mathbb{Z}_{k, i}) \oplus \mathbb{W}_k
\end{equation}

With that we can propose the following zonotope propagation law:
\begin{equation}
\label{eq:LTV_zonotopes_robust_propagation}
    \mathbb{X}_{k+1} = 
    \text{Co}(
    A_i(k) \mathbb{X}_k + B_i(k) \mathbb{U}_k + d_i(k) )
    \oplus \mathbb{W}_k,
\end{equation}

Now we observe the reason for separating $d(k)$ and $\mathbb{W}_k$; if they were combined in $\Zo{d(k)}{W(k)}$, then a different Minkowski sum would be applied to each vertex model, leading to a significant increase in the order of the resulting zonotopes.

\subsection{Matrix zonotope-based propagation}

Assuming that $d(k)$ is known exactly and set $\Omega_k$ can be represented as matrix zonotopes $\mathbb{A}(k)$ and $\mathbb{B}(k)$, such that 
\begin{equation}
    [A(k), B(k)] \in \Omega_k \iff A(k) \in \mathbb{A}(k), B(k) \in \mathbb{B}(k)
\end{equation}

Both the assumption on $d(k)$ and on $\Omega$ are restrictive. However, they allow an alternative formulation of zonotope propagation. First, we relax \eqref{eq:LTV_zonotopes_robust} as:
\begin{equation}
\label{eq:LTV_zonotopes_robust_matrix_1}
    \mathbb{X}_{k+1} \supset 
    \mathbb{W}_k \oplus 
    \bigcup\limits_{\Omega_k} 
    \left( A(k) \mathbb{X}_k + B(k) \mathbb{U}_k + d(k) \right)
\end{equation}

Obtained expression \eqref{eq:LTV_zonotopes_robust_matrix_1} can be over-approximated using matrix zonotope multiplication \eqref{eq_matrix_zonotope_mult}:
\begin{equation}
\label{eq:LTV_zonotopes_robust_propagation_matrix}
    \mathbb{X}_{k+1} = 
    \left( \mathbb{A}(k) \otimes \mathbb{X}_k + \mathbb{B}(k) \otimes \mathbb{U}_k + d(k) \right) \oplus \mathbb{W}_k 
\end{equation}

Same as the formulation \eqref{eq:LTV_zonotopes_robust_propagation}, this propagation law leads to rapid growth in the zonotope order, requiring order reduction to be applied.

\subsection{Zonotope Order Reduction}

Methods proposed in this work lead to a steady increase in the zonotope order at consecutive time steps, due to the use of Minkowski sum and approximate convex hull operations. Not only is it preferable to maintain zonotope order uniform, but its increase leads to a higher number of decision variables in the resulting optimal control problem (OCP). This issue has been well-recognized in the literature \cite{kopetzki2017methods, raghuraman2020set, yang2018comparison}. It is usually mitigated with order reduction methods.

A number of order reduction methods have been previously proposed, including ones based on SVD, exhaustive search, sorting, and various non-convex procedures. Since our goal is to solve the control design problem as a single convex program, we seek to embed the order reduction method in it. Moreover, since this convex program includes an iterative application of the previously mentioned Minkowski sum and approximate convex hull operations, the desirable order reduction method should be suitable for iterative application in the same manner.

We propose \emph{ReaZOR} (\textbf{Rea}rranging \textbf{Z}onotope \textbf{O}rder \textbf{R}eduction) method, that operates by shuffling zonotope generator vectors and replacing a subset of those vectors with a smaller set, such that the resulting zonotope includes the original one but has a smaller order. ReaZOR takes as an input a generator $G = (g^{(1)}, ..., g^{(z)}) \in \mathbb{R}^{n \times z}$, and outputs an order-reduced generator $G_{red} \in \mathbb{R}^{n \times p}$:

\begin{equation}
\label{eq:ReaZOR}
    \begin{aligned}
    G_{\text{red}} & = & \underset{G^*, a_i}{\text{argmin}} &  \ 
    \sum\limits_{i=1}^n  |a_i| \\[0.5em]
            &   & \text{s. t.} &  
            \ \sum\limits_{j=p-n+1}^z  |G_{ij}| \leq a_i, \ i = 1, ..., n \\
            &   &   &  \ G^* = 
            \Bigl(
            \text{diag}(a_1, ..., a_n), g^{(1)},...,g^{(p-n)}
            \Bigr)
    \end{aligned}
\end{equation}
where $a_i \in \mathbb{R}$, $n \leq p \leq z$. 

the key idea of this algorithm is that row-wise approximation is applied to the last columns of the generator but the diagonal matrix resulting from this approximation is placed as the first $n$ columns of the reduced generator; the first $p-n$ columns of the old generator are pushed to the back of the new one. This operation achieves greater uniformity in column lengths and makes the algorithm numerically stabler. The only hyperparameter in the algorithm is the number of columns $p$ in the reduced zonotope In our experiments we found that the resulting optimization program is reasonably sensitive to the choice of this parameter, as is to be expected. 

ReaZOR is designed to be applied iteratively and as a part of a convex optimization problem. As such it does not compete with order reduction methods based on non-linear operations. We can illustrate its behavior with the following example:

\begin{example}
\label{example_ReaZOR}
Apply ReaZOR to zonotope $\Zo{0}{G}$: 
\[G = 
\begin{bmatrix}
    4 & 2 & 2 & 1 & 1 \\
    4 & 1 & 0 & 2 & 1 \\
\end{bmatrix}, 
\]
with number of columns $p = 4$ after reduction. We obtain $a_1 = 4$ and $a_2 = 3$, and the resulting zonotope $\Zo{0}{G_{\text{red}}}$ has the following generator:
\[
G_{\text{red}} = 
\begin{bmatrix}
    4 & 0 & 4 & 2 \\
    0 & 3 & 4 & 1 \\
\end{bmatrix}.
\]
To illustrate behavior of the algorithm under iterative application, we find Minkowski sum $\Zo{0}{G_{\text{red}}} \oplus \Zo{0}{\begin{bmatrix}
    1 \\ 1
\end{bmatrix}}$ and apply ReaZOR to the result:
\begin{align*}
\text{ReaZOR}\left( 
\begin{bmatrix}
    4 & 0 & 4 & 2 & 1 \\
    0 & 3 & 4 & 1 & 1 \\
\end{bmatrix} 
\right) =
\begin{bmatrix}
    7 & 0 & 4 & 0  \\
    0 & 6 & 0 & 3  \\
\end{bmatrix}.
\end{align*}
\end{example}

As illustrated by the example \ref{example_ReaZOR}, ReaZOR creates and shuffles diagonal blocks in the zonotope generator. If $W(k)$ is diagonal, ReaZOR allows us to use the property \eqref{diagMinkowski} to perform Minkowski addition without the increase of the zonotope order.

\subsection{Cost design}

Cost design is an important problem for optimization-based methods. Here we propose a three-component cost:
\begin{equation}
    J = J_c + J_g + J_r
\end{equation}
where $J_c$ is the cost on deviation from the nominal trajectory  applied to the zonotope centers $\Bar{x}_k$ and $\Bar{u}_k$; $J_g$ is the cost on zonotope size applied to the generators $G_k$ and $\theta_k$, and $J_r$ is the cost associated with the order reduction algorithm \eqref{eq:ReaZOR}:
\begin{equation}
\begin{aligned}
    J_c = \sum_k^N \left(
                 (\Bar{x}_k - x^*_k)^\top Q_c (\Bar{x}_k - x^*_k) \right) + \\
          \sum_k^{N-1} \left(
                 (\Bar{u}_k - u^*_k)^\top R_c (\Bar{u}_k - u^*_k) \right), \\
    J_g = \sum_k^N \text{Tr} 
                 \left(G_k Q_g G_k^\top \right) 
                 + 
                 \sum_k^{N-1} \text{Tr} 
                 \left(\theta_k R_g \theta_k^\top
                 \right).             
\end{aligned}
\end{equation}
where $Q_c$, $R_c$, $Q_g$ and $R_g$ are positive-definite weight matrices and $\text{Tr}(\cdot)$ is a trace operation.

\subsection{Control design as a convex program}

Combining robust propagation with order reduction, we can formulate the following OCP:
\begin{equation} \label{ConvexOptimization}
\begin{aligned}
& \{\mathbb{X}_{k}, \mathbb{U}_k\} \ = \ {\text{argmin}} \ J, \\
& \text{s. t.} \ 
\begin{cases}
\mathbb{X}^*_k = 
\underset{\mathcal{V}_k}{\text{Co}}
( A_i(k) \mathbb{X}_k + B_i(k) \mathbb{U}_k + d_i(k)), \\ 
\mathbb{X}_{k+1} = \text{ReaZOR}(\mathbb{X}^*_k) \oplus \mathbb{W}_k, \\
\mathbb{X}_{N} \subseteq \mathbb{H}_g^x, \mathbb{X}_{k} \subseteq \mathbb{H}_k^x, \mathbb{U}_k \subseteq \mathbb{H}_k^u, \\
[A_i, B_i, d_i] \in \mathcal{V}_k, \\
k \in \{1,...,N-1\}
\end{cases}
\end{aligned}
\end{equation}
where zonotope inclusion constraints are implemented using linear constraints \eqref{eq:containment}, and $\text{ReaZOR}()$ refers to the inclusion of the constraints and cost from \eqref{eq:ReaZOR}; $\mathbb{H}_g^x$ is a zonotope that represents bounds on the final state in the trajectory. Since the cost is a positive-definite quadratic function, all equality constraints are linear, and inequality constraints are either linear or conic (for order reduction), the problem is convex. Decision variables in this problem are $\mathbb{X}_{k}$, $\mathbb{U}_k$, and $\mathbb{X}^*_k$.

\subsection{Hybrid dynamics}

Assume we have a hybrid dynamical system, described as follows:

\begin{equation}
\label{eq:hybridLTV}
    \begin{matrix}
    x_{k+1}  =  A(k) x_k + B(k) u_k + d(k) + w_k,
    \\[0.5em]
    [A(k) \ B(k) \ d(k)] \in \Omega_k^j, \ \ \ x_k \in \mathbb{H}_j
    \\[0.5em]
    \end{matrix}
\end{equation}
where state-space is divided into non-intersecting regions $\mathbb{H}_j$ (where $j \in [1, p]$), and each region has associated uncertain time-varying linear dynamical model $[A(k) \ B(k) \ d(k)] \in \Omega_k^j$; when the true state of the system changes from one region to another, the dynamics switches. Assuming that regions $\mathbb{H}_j$ 
 are described as zonotopes $\mathbb{H}_j = \langle h^x_j, H^x_j \rangle$ and introducing binary variables $c_{k, j}$ we can write the hybrid version of the OCP proposed in the previous subsection:
\begin{equation} \label{ConvexOptimization_hybrid}
\begin{aligned}
& \{\mathbb{X}_{k}, \mathbb{U}_k\} \ = \ {\text{argmin}} \ J, \\
& \text{s. t.} \ 
\begin{cases}
\mathbb{X}^*_{k, j} = 
\underset{\mathcal{V}_{k, j}}{\text{Co}}
( A_i(k) \mathbb{X}_k + B_i(k) \mathbb{U}_k + d_i(k)), \\ 
||\mathbb{X}^*_k - \mathbb{X}^*_{k, j}||_F \leq M (1 - c_{k, j}), \\ 
\mathbb{X}_{k+1} = \text{ReaZOR}(\mathbb{X}^*_k) \oplus \mathbb{W}_k, 
\\
G_{k} = H^x_j \Gamma_{k, j}, \ \ 
h^x_j - \bar{x}_{k} = H^x_j \beta_{k, j}, \\ 
||(\Gamma_{k, j},\beta_{k, j})||_\infty \leq 1 + M (1 - c_{k, j}), 
\\
\mathbb{X}_{N} \subseteq \mathbb{H}_g^x, \ \mathbb{U}_k \subseteq \mathbb{H}_k^u, \\
\sum\limits_{j=1}^p c_{k, j} = 1, \\
c_{k, j} \in \{0, \ 1 \}, \ k \in \{1,...,N-1\}, \ j \in \{1,...,p\}
\end{cases}
\end{aligned}
\end{equation}
where $\Gamma_{k, j}$ and $\beta_{k, j}$ are containment coefficients (see eq. \eqref{eq:containment}), $M$ is a sufficiently large constant, $c_{k, j}$ are binary variables, implementing choice between hybrid dynamic modes, and zonotope norm $||\cdot||_F$ is Frobenius norm of zonotope generator concatenated with zonotope center $|| \langle c, G \rangle ||_F = || (c, G) ||_F$. In this formulation the same set of binary variables $c_{k, j}$ links the region $\mathbb{H}_j$ with the associated dynamics $\mathcal{V}_{k, j}$. Let us note that the computational time for mixed integer problems grows with the number of integer variables, which in this case depends on the number of propagation steps and on the number of hybrid modes \cite{sadraddini2019sampling}.

\subsection{Finding current zonotope and policy}
\begin{figure}[t]
	\centering
	\includegraphics[ width=0.9\columnwidth]{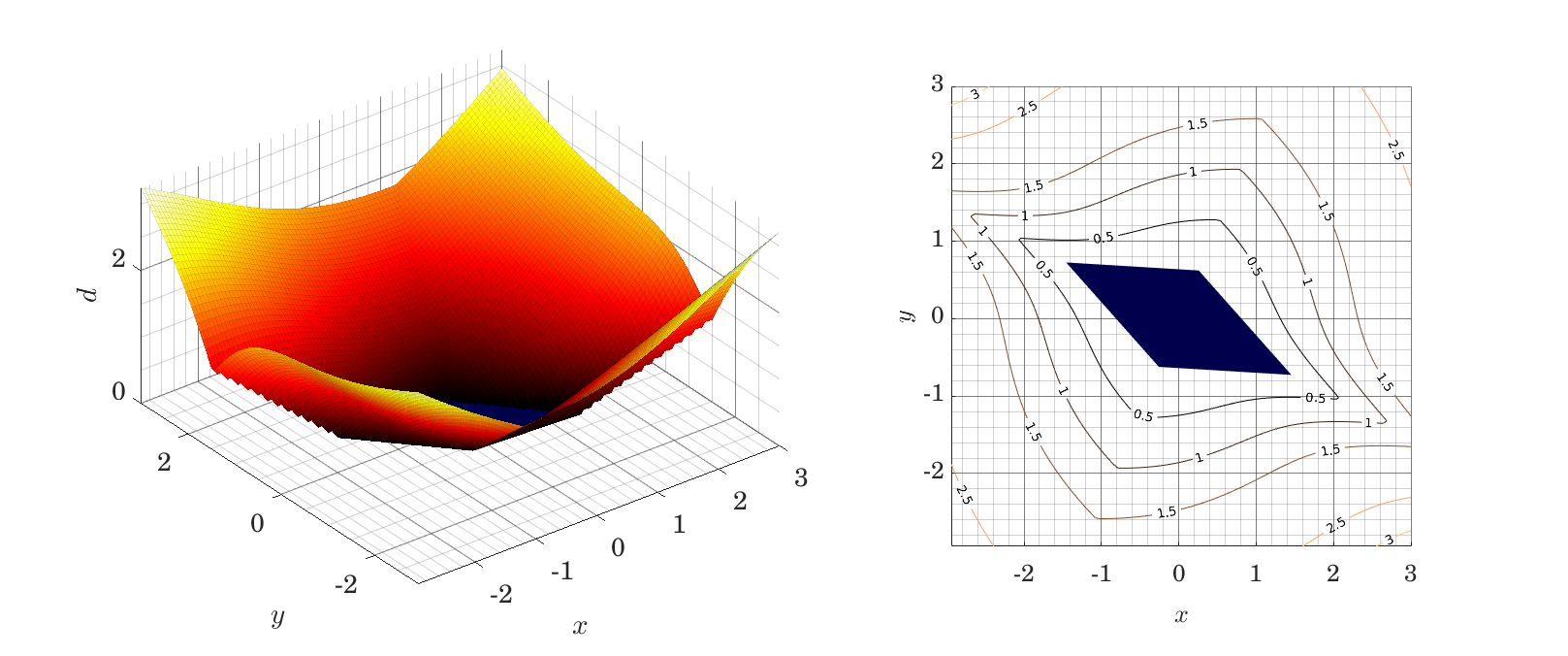}
	\caption{Distance to parallelotope function representation as a surface and a contour map}
	\label{fig:distance_3d}
\end{figure}

It is often meaningful to draw a distinction between control policies that depend purely on the state of the system, and the ones that depend on time as well. In the field of orbital stabilization, it is common to use transverse dynamics formulations to write the control in a time-free form \cite{manchester2011transverse, shiriaev2008can}. In the explicit MPC \cite{bemporad2000piecewise, alessio2009survey} and in the tree-based algorithms \cite{sadraddini2019sampling, tedrake2010lqr}, control policies can be selected based on the region of the state space the system is currently in. This method becomes challenging in implementation when a tessellation of the state-space is not available, and instead, the state-space is only partly covered by overlapping regions, which is the expected situation when zonotopes or ellipsoids are used for set representation. Additionally, discrete switching between control policies may lead to chattering. Finally, we observe that zonotope propagation in eq. \eqref{eq:LTV_zonotopes} naturally leads to a one-to-one correspondence between the sets $\mathbb{X}_k$ and $\mathbb{U}_k$. However, it does not yield a linear control law, which means that control law \eqref{eq:lin_feedback} does lead to the exact execution of the designed propagation. With that in mind, if the current state of the system $x$ belongs to several zonotopes, it is preferable to avoid the ones whose boundary is close to the state $x$. Thus, we propose a quasi-time-free control policy choice algorithm.

We distinguish two basic scenarios with regard to the current state of the system and policy choice: 1) the current state $x$ lies in one or more zonotopes $\mathbb{X}_k$ and 2) $x$ does not lie in any zonotope $\mathbb{X}_k$. In the first case, the problem is to choose which zonotope $\mathbb{X}_k \ni x$ to use for policy generation; in the second case, the problem is to find the nearest zonotope to $x$ and apply its control policy. This problem has previously been studied in \cite{sadraddini2019sampling, wu2020nearest}. 

We propose the following quasi-time-free solution to the first case: we register the \emph{sequence number} $k$ of last zonotope $\mathbb{X}_k$, whose control policy \eqref{eq:lin_feedback} was applied. Then if $x \in \mathbb{X}_{k+1}$, we apply the control policy associated with the $\mathbb{X}_{k+1}$ zonotope; if not, we find among $\mathbb{X}_k \ni x$ zonotope whose center is the nearest to $x$ in the Euclidean sense and apply the control policy associated with it. The priority given to $\mathbb{X}_{k+1}$ zonotope avoids the effect of chattering; the use of the zonotope with the nearest center partially avoids the problems resulting from the linear approximation of the control policy map discussed above.

In order to propose a solution to the second case, we need to provide a distance-to-zonotope function. This function will be running in real-time and therefore is required to be computationally light. With that in mind, we propose an additional offline step of computing parallelotope bounds $\mathbb{P}_{k}$ for each zonotope $\mathbb{X}_{k}$, using PCA-based algorithm reported in \cite{kopetzki2017methods}. Then we can use the following vector-to-zonotope distance function:
\begin{align}
    & d(x, x_c, P) = \nu \cdot \text{max}\{0, \ \norm{P^+ (x - x_c)}_\infty - 1\}  \\
    & \nu(x, x_c, P) = \frac{\norm{x - x_c}}{\norm{P^+ (x - x_c)}}
\end{align}
where $P$ is the generator of the parallelotope $\mathbb{P}$, $x_c$ is the center of $\mathbb{P}$ and $\mathbb{X}$, and $\nu(x, x_c, P)$ is a scaling factor.
%

\input{sections/Algorithm}

The resulting control policy choice is made with the algorithm~\ref{alg:ContolLaw}. To speed it up, we use a k-d tree algorithm to choose $n$ closest zonotopes and run algorithm~\ref{alg:ContolLaw} on them.

\begin{figure}
	\centering
	\includegraphics[ width=0.3\columnwidth]{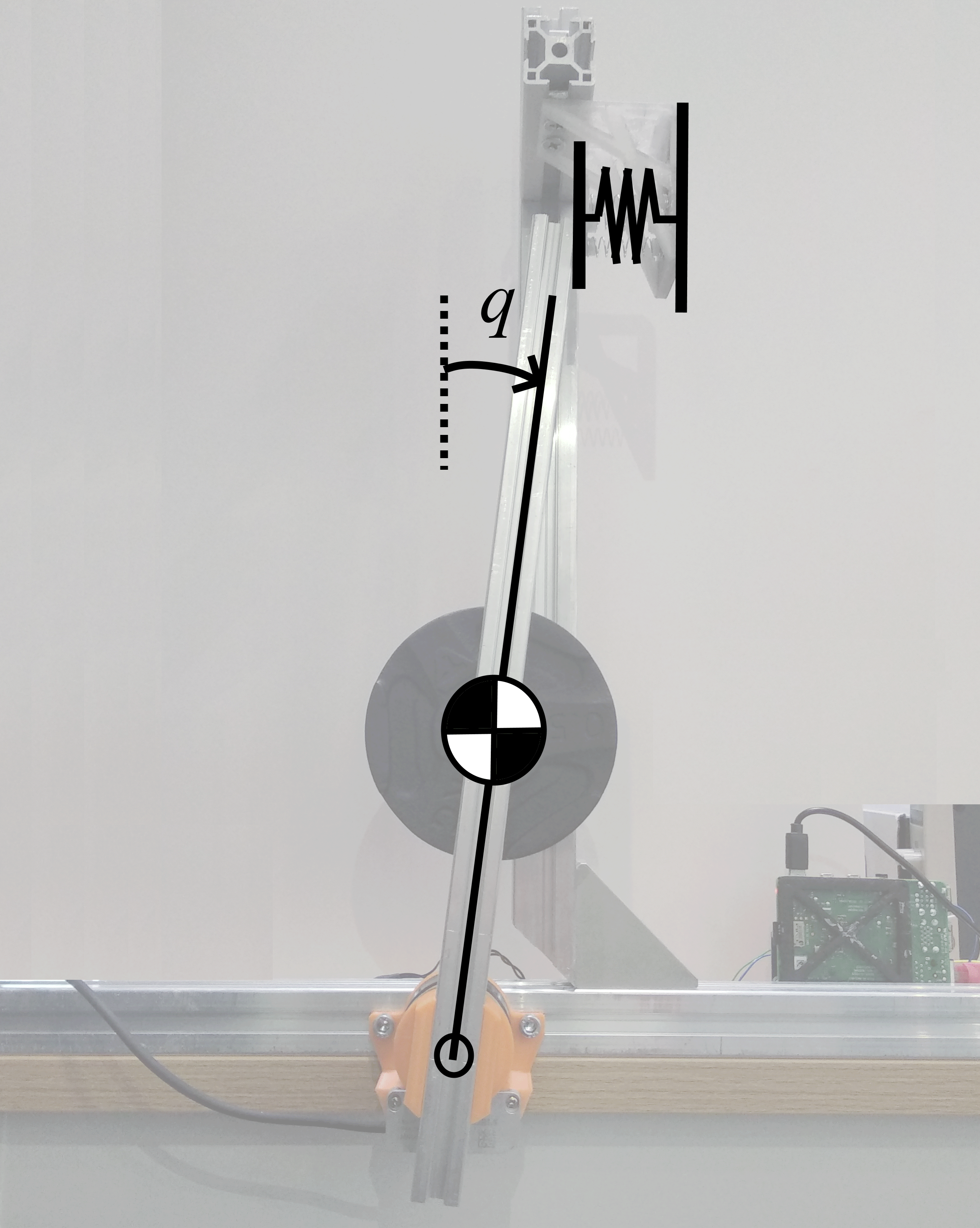}
	\caption{Experimental setup: Inverted pendulum with an elastic wall; $q$ is the angle between the vertical line and the pendulum's shaft}
	\label{fig:pendulum}
\end{figure}

The next sections demonstrate experimental validation of the proposed algorithm on two experimental set-ups: \emph{inverted pendulum with a wall} and \emph{pendubot}.

%% file: sections/Algorithm.tex
%

\begin{algorithm}
\caption{Time-free control policy choice}
\label{alg:ContolLaw}
\begin{algorithmic}
    \State \textbf{Data: } $x$, $P_i$, $G_i$, $\bar x_i$, $\theta_i$, $\bar{u}_i$, $k$
    \State \textbf{Result: } $u$
    \State $d_i := d(x, x_{c, i}, P_i), \ \forall i$\;
    \If {$d_{k+1} = 0$}{}
        \State $\kappa := k+1$
    \Else 
        \If {$\exists j, \ d_j = 0$}{}
            \State $\kappa := \underset{j}{\text{argmin}}\{||x - \bar x_j||: \ d_j = 0 \}$
        \Else
            \State $\kappa := \underset{j}{\text{argmin}} \ |d_j| $
    \EndIf
    \EndIf
    \State $u = \bar{u}_\kappa - \theta_\kappa G_\kappa^\dagger(x - \bar x_\kappa)$
\end{algorithmic}
\end{algorithm}

%% file: sections/sect_5_pendulum.tex
\section{Case-Study: Inverted Pendulum with a Wall} \label{sec_InvertedPendulum}

For the validation of the proposed methods, we take an example of a hybrid system - an inverted pendulum that interacts with an elastic wall. This system has been used for a similar purpose in \cite{sadraddini2019sampling,marcucci2017approximate}.

\subsection{Experimental setup}
Our setup consists of an 18V BLDC motor GYEMS RMD-L-50 and its driver GYEMS DRC-06, allowing current control; we use Renishaw MHA7 for position measurements and CAN BUS for communications, with the control commands updated on 250Hz. The program is being run on Raspberry PI 4 ModelB. A 20x20x410mm aluminum profile is attached to the shaft of the motor; the profile has sockets for attaching an additional mass at different distances from the motor shaft.

\subsection{Mathematical description}

The position of the pendulum is described by angle $q$. The dynamics of the system has two hybrid states: with contact ($q \geq q_c$) and without it ($q < q_c$), where $q_c$ is the angle at which the contact with the undeformed wall takes place:

\begin{equation*}
\begin{cases}
        I \ddot q + \mu_f \dot q + m g l \sin(q) = c_{\tau} i & \text{if} \ q < q_c \\
        I \ddot q + \mu_c \dot q + m g l \sin(q) + k(q - q_c)= c_{\tau} i & \text{if} \ q \geq q_c \\
\end{cases}
\end{equation*}
where $I$, $m$, $l$ are the moment of inertia, mass, and length of the pendulum, $\mu_f$ and $\mu_c$ are viscous friction forces, $g$ is the gravitational constant, $c_{\tau}$ is the torque coefficient, and $k$ is the stiffness coefficient of the wall.

\begin{figure}[h]
	\centering
	\includegraphics[width=0.5\textwidth]{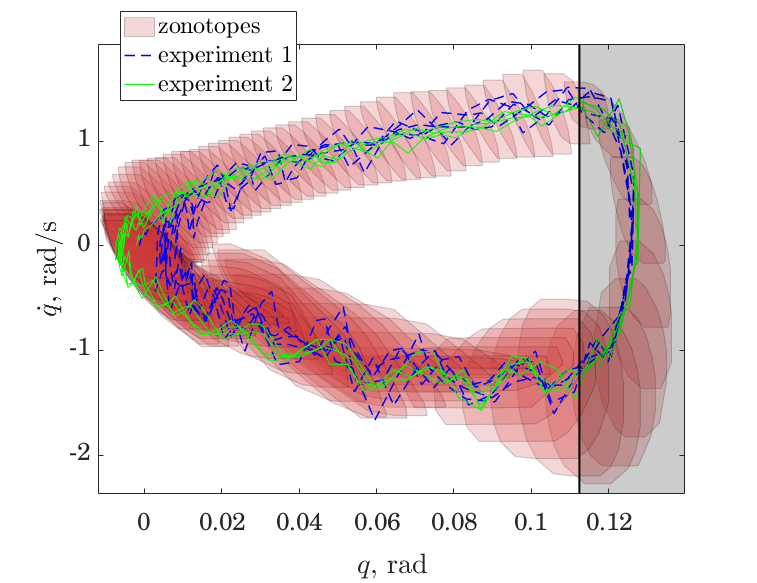}
	\caption{Zonotope propagation, obtained via proposed method, and state-space trajectories of the pendulumn, recorded during two experiments with different values of $l$; grey area denotes states in collusion with the elastic wall.}
	\label{fig:phase_pendulum}
\end{figure}

Some of the parameters are known exactly: $m = 0.126$ kg,  $\mu_f = 0.001$ N/ms, $c_{\tau} = 0.03$ Nm/A. Others are known to be in the intervals: $I \in [0.0116, 0.0203] \ \text{m}^2$, $k \in [116.1, 141.9]$ N/rad, $\mu_c \in [0.41, 0.51]$ N/ms and $l \in [0.12, 0.18]$ m. This allows us to define two sets of vertices $\mathcal{V}_{k, 1}$ for the case $q \geq q_c$, containing variations of parameters $I$ and $l$ and $\mathcal{V}_{k, 2}$ for the case $q < q_c$, containing variations of parameters $k$ and $\mu_c$, with the assumption that when the pendulum touches the wall, these two parameters dominate its dynamics. In both cases $\mathcal{V}_{k, i}$ will have four elements.

Initial and final sets $\mathbb{X}_0$ and $\mathbb{X}_f$ are constrained as: $\mathbb{X}_0 = \mathbb{X}_f \in \Zo{0_{2 \times 1}}{\text{diag}(0.02, 0.4)}$, where $\text{diag}$ is an operator that returns a matrix with its inputs on the diagonal, and $0_{2 \times 1}$ is a vector of zeros. Additive disturbance $\mathbb{W}$ is chosen as $\mathbb{W} = \Zo{0_{2 \times 1}}{\text{diag}(10^{-4}, 10^{-3})}$, torque limits are set implicitly as $\mathbb{H}^u = \Zo{0}{20}$. Parameter $z$ (zonotope order) in the order reduction algorithm is chosen as 6.

The control law was applied to the experimental setup. The first experiment was performed for $ l = 0.13 $ m, the second for $ l = 0.17 $ m. Results of the control design and the experiments are shown together in Fig. \ref{fig:phase_pendulum}. It was possible to successfully design a control policy by using proposed methods as well as verify its performance on the experimental setup.

%% file: sections/sect_6_pendubot.tex
\section{Case-Study: Pendubot}
\begin{figure}[t]
	\centering
	\includegraphics[width=0.25\textwidth]{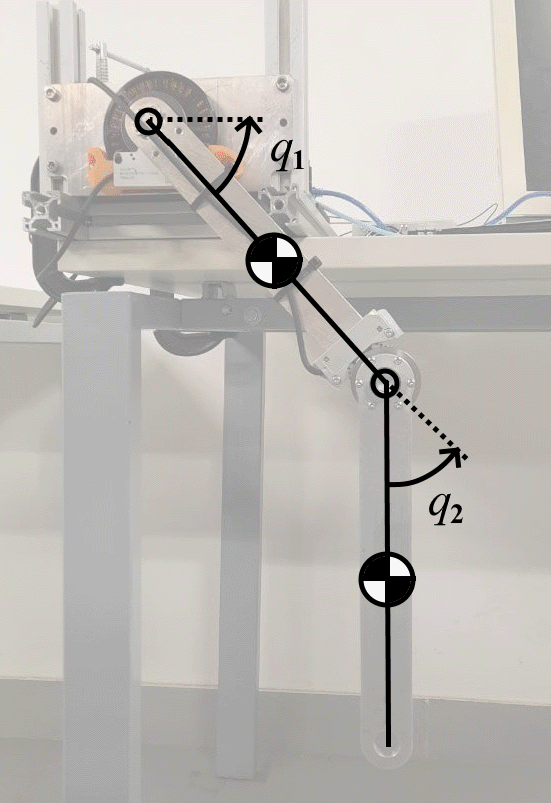}
	\caption{Experimental setup: a pendubot robot; $q_1$ and $q_2$ are joint angles}
	\label{fig:pendubot}
\end{figure}
We would like to demonstrate the performance of our algorithm for an underactuated system performing a non-trivial motion. To demonstrate this, we will take pendubot, a classic example of an underactuated system discussed in \cite{freidovich2008periodic, fantoni2000energy}
\subsection{Experimental setup}
Pendubot is a two-link planar manipulator with an actuator in the first link and a passive second link. We use a BLDC motor (T-Motor U8 Lite KV85) with an ODrive controller, allowing current control. Positions of the links are measured with Renishaw MHA7 encoders. Control commands are updated at the 100Hz frequency. A diagram of the robot is shown in Fig. \ref{fig:pendubot}
\subsection{Mathematical description}
\begin{figure}[bp]
	\centering
	\includegraphics[width=0.5\textwidth]{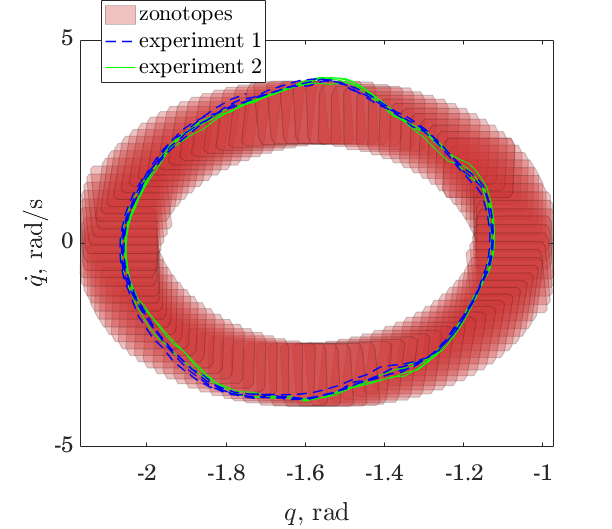}
	\caption{Zonotope propagation (in projection on the $q_1, \dot q_1$ subspace), obtained via the proposed method, and state-space trajectories of the pendulum, recorded during two experiments with different values of $m_e$}
	\label{fig:phase_pendubot}
\end{figure}
Pendubot dynamics can be described in the following form:
\begin{equation}\label{eq:pendubot_dynamics}
D(q) \ddot q + C(q,\dot q) \dot q + g(q) = B u - f_f
\end{equation}	
where $q = [q_1,q_2]^\top$ defines orientation of the links, $D$ is the generalized inertia matrix, $C$ is the Cariolis and inertial force matrix, $B$ is the control matrix, and $f_f$ is the generalized bearing friction. Except for the bearing friction, analytical expressions for these quantities can be found in the literature \cite{freidovich2008periodic, fantoni2000energy}. Bearing friction can be described as:
\begin{equation}
    f_f = b_s\text{sgn}(\dot q) + b_v \dot q
\end{equation}
where $b_s$ and $b_v$ are constants and $\text{sgn}$ is element-wise sign function. 

The lengths and masses of the system parts were measured directly, the moments of inertia were taken from the design documentation, and the friction coefficients were obtained by identification. However, to demonstrate the robustness of the algorithm, in some experiments, an additional mass $m_e = 0.06$ kg is attached to the middle of the first link.

The trajectory of the system was obtained via direct collocation as a solution to non-convex optimization, followed by linearization and discretization along the found trajectory. Initial and final sets $\mathbb{X}_0$ and $\mathbb{X}_f$ are constrained as: $\mathbb{X}_0 = \mathbb{X}_f \in \Zo{0_{4 \times 1}}{\text{diag}(0.2, 0.2, 2, 2)}$. Additive disturbance $\mathbb{W}$ is chosen as $\mathbb{W} = \Zo{0_{4 \times 1}}{9.5 \cdot 10^{-4} I}$, where $I$ is identity matrix. Torque limits are set implicitly as $\mathbb{H}^u = \Zo{0}{10}$. Parameter $z$ (zonotope order) in the order reduction algorithm is chosen as 62.5, much higher than in the previous example.

The control law was applied to the experimental setup. The first experiment was performed with additional mass $m_e = 0.06$ kg on the first link, and the second - without the additional mass. Results of the control design and the experiments are shown together in Fig. \ref{fig:phase_pendubot}. The method allowed us to design a control law that was able to stabilize the trajectory, which was shown both in simulation and via experimental study.

Control design for this and previous experiments can be replicated using our code, distributed under open source license \cite{Balakhnov2021}.

%% file: sections/sect_7_order_reduction.tex
\section{Comparative study of order reduction methods}
\begin{figure}[t]
	\centering
	\includegraphics[width=0.5\textwidth]{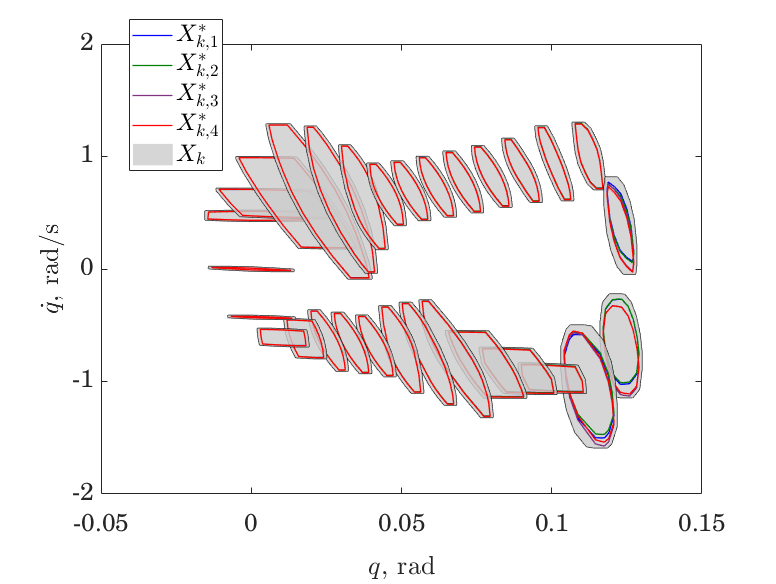}
	\caption{Comparison of order reduction methods and their influence on zonotope volume; zonotope propagation illustrated here is obtained via the proposed method, using state-space trajectories of the pendulumn with a wall model (only 1 in 3 consecutive zonotopes shown). Zonotopes $\mathbb{X}_{k,i}^*$ are drawn by colored outlines, zonotope $\mathbb{X}_k$ is drawn in grey.}
	\label{fig:convhull_reduction}
\end{figure}  
In this section, we study the conservativeness of the proposed zonotope order reduction method ReaZOR with respect to other known order reduction methods. While other methods can not replace ReaZOR as a part of a convex optimization program, it is still important to analyze its performance. We compare it with order reduction methods implemented in software package CORA \cite{althoff2015introduction, CORA2020}: 1) Girard's method proposed in \cite{girard2005reachability}, 2) Combastel's method proposed in \cite{combastel2003state}, 3-4) volume minimization methods (denoted as Method A and Method C in the original paper) \cite{althoff2010reachability}, 5) method proposed by Scott et. al. \cite{scott2016constrained}, and
6) principal component analysis (PCA)-based method reported in \cite{kopetzki2017methods}.
        
The comparative study is performed as follows. We take a control design solution for inverted pendulum with a wall discussed in the Section \ref{sec_InvertedPendulum} represented by a sequence of zonotopes $\mathbb{X}_k$ and $\mathbb{U}_k$. To each zonotope $\mathbb{X}_k$ we apply propagation $\mathbb{X}_{k,i}^* = A_i(k) \mathbb{X}_k + B_i(k) \mathbb{U}_k + d(k)$ individually for all $[A_i, B_i, d_i] \in \mathcal{V}_k$. Then we find overapproximation \eqref{eq_convhull} of the convex hull of resulting zonotopes $\mathbb{X}_{k,i}^*$, denoted as $\mathbb{X}_k^*$. To follow algorithm \eqref{ConvexOptimization} we need to perform order reduction on $\mathbb{X}_k^*$ and apply Minkowski sum to it, thus finding $\mathbb{X}_k$.

Zonotope $\mathbb{X}_k^*$ has $r$ columns, and we reduce it to $p$ columns using all methods listed above. Let $V_{k,h}$ be the volume of the zonotope $\mathbb{X}_k^*$ after reduction by method \#$h$, where the first six methods have been listed above, ReaZOR is the method number 7; $V_{k,8}$ corresponds to the volume of $\mathbb{X}_k^*$ without reduction. Let $\nu_{k}$ be the percentage difference between the volume of a reduced zonotope and the volume of the zonotope without reduction $V_{k,8}$:
\begin{equation}
    \nu_{k, h} = \frac{| V_{k,h} - V_{k,8} |}{V_{k,8}} \cdot 100\%
\end{equation}
%
We will refer to $\nu_{k, h}$ as volume errors. In this experiment, we use zonotope propagation for the uncertain hybrid dynamics described in Section \ref{sec_InvertedPendulum}; the resulting sequence of shown in Figure \ref{fig:convhull_reduction}. The goal of the next experiment is to show how various order reduction techniques differ in terms of volume error. 
\begin{table}[t]
\caption{Comparison of mean and maximum values of volume errors of the reduced zonotopes for different order reduction methods}
\label{table_experiment1}
\begin{center}
\begin{tabular}{|c|c|c|c|c|}
\hline
\# & Method & Reference &  $\underset{k}{\text{mean}}( \nu_{k, h} )$ & $\underset{k}{\text{max}}( \nu_{k, h} )$ \\
\hline
1 & Girard's & \cite{girard2005reachability, CORA2020} & $0.673$\% & $1.995$\% \\
\hline
2 & Combastel's & \cite{combastel2003state, CORA2020} & $0.673$\% & $1.995$\% \\
\hline
3 & Method A & \cite{althoff2010reachability, CORA2020} & $0.673$\% & $1.995$\% \\
\hline
4 & Method C & \cite{althoff2010reachability, CORA2020} & $0.673$\% & $1.995$\% \\
\hline
5 & Scott et. al. & \cite{scott2016constrained, CORA2020} & $0.674$\% & $1.995$\% \\
\hline
6 & PCA-based & \cite{kopetzki2017methods, CORA2020} & $0.673$\% & $1.995$\% \\
\hline
7 & ReaZOR & & $0.675$\% & $1.981$\% \\
\hline
\end{tabular}
\end{center}
\end{table}
Table \ref{table_experiment1} shows the mean and maximum values of $\nu_{k, h}$ across the whole trajectory for each method included in the comparison. As we can see, ReaZOR is slightly better than the other methods in terms of the maximum value of volume error and is slightly worse than the others in terms of the mean value of volume error; but in both cases, the differences are negligible compared with the errors themselves. This indicates that the volume errors are dominated by the geometry of the sets rather than by particular features of the order reduction methods. This allows us to conjecture that in terms of volume error, all methods show very similar performance, meaning that other metrics can be used in choosing the preferred order reduction method; in our case, we value numerical properties that the methods exhibit when used as a part of a convex optimization problem.

%% file: sections/sect_8_conc.tex
\section{Conclusions}

In this paper, we proposed and experimentally validated zonotope-based robust explicit MPC, as well as additional algorithms: convex order reduction method and time-free policy choice. It was experimentally shown that the methods work on hybrid systems, non-linear systems linearized along a trajectory, systems with parametric uncertainty in friction, mass, inertia, and stiffness.

The proposed method can be used to account for a variety of sources of parametric uncertainty, including linearization errors. The method is conservative by design; we observed that even when the system exits state-space areas covered by zonotopes (which happens due to poor sensory feedback, large unmodelled disturbances, linearization errors, or unmodelled dynamics), it still tends to return to one of the zonotopes. A further study of this property might be of interest.